\pdfoutput=1

\documentclass[11pt]{article}

\usepackage[final]{acl}

\usepackage{times}
\usepackage{latexsym}
\usepackage{booktabs}
\usepackage{makecell}
\usepackage{tabularx} 
\usepackage{geometry} 
\usepackage{caption}
\usepackage{amsmath}
\usepackage{amsfonts}
\usepackage{pdfpages}
\usepackage{fancyhdr}
\usepackage{graphicx} 
\usepackage{float}
\usepackage{subfigure}
\newcolumntype{Y}{>{\centering\arraybackslash}X}
\newcolumntype{m}{>{\hsize=.05\hsize}X}
\newcolumntype{H}{>{\setbox0=\hbox\bgroup}c<{\egroup}@{}}

\usepackage[T1]{fontenc}

\usepackage[utf8]{inputenc}

\usepackage{microtype}

\usepackage{inconsolata}

\usepackage{graphicx}

\usepackage{xcolor,colortbl}

\usepackage{tcolorbox}
\tcbuselibrary{theorems}
\newtcbtheorem[]{mytheo}{Prompt}%
{colback=green!5,colframe=green!35!black,fonttitle=\bfseries}{th}

\title{ArithmAttack: Evaluating Robustness of LLMs to Noisy Context \\in Math Problem Solving}

\author{Zain Ul Abedin\textsuperscript{$\clubsuit$}\thanks{Equal contribution} \hspace{3mm}  Shahzeb Qamar\textsuperscript{$\clubsuit$}\footnotemark[1] \hspace{3mm} Lucie Flek\textsuperscript{$\clubsuit\spadesuit$} \hspace{3mm} Akbar Karimi\textsuperscript{$\clubsuit\spadesuit$}\\
\vspace{-1mm}\\
\textsuperscript{$\clubsuit$}Bonn-Aachen International Center for Information Technology, University of Bonn, Germany\\
\textsuperscript{$\spadesuit$}Lamarr Institute for ML and AI, Germany\\
\texttt{ak@bit.uni-bonn.de}}

\begin{document}

\maketitle

\begin{abstract}
While Large Language Models (LLMs) have shown impressive capabilities in math problem-solving tasks, their robustness to noisy inputs is not well-studied. We propose \textbf{ArithmAttack} to examine how robust the LLMs are when they encounter noisy prompts that contain extra noise in the form of punctuation marks. While being easy to implement, ArithmAttack does not cause any information loss since words are not added or deleted from the context. We evaluate the robustness of eight LLMs, including LLama3, Mistral, Mathstral, and DeepSeek on noisy GSM8K and MultiArith datasets. Our experiments suggest that all the studied models show vulnerability to such noise, with more noise leading to poorer performances.
\end{abstract}

\section{Introduction}

As Large Language Models (LLMs) are improving in their ability to accurately process human language, their math problem-solving is also enhancing \cite{saraf2024towards, agrawal24give, wu2024mathchat}. However, these sets of questions might require reasoning capabilities to be answered. While LLMs have been shown to have such capabilities to some extent \cite{imani2023mathprompter}, their robustness to adversarial inputs remains a challenge. For instance, these models can be vulnerable to simple replacement of words with their synonyms \cite{zhou2024mathattack} and even typographical errors can negatively impact their ability to reason \cite{gan2024reasoning}. However, such attacks can semantically alter the samples by changing the current sample features to completely different ones (e.g. amoral --> moral).

\begin{figure}
    \centering
    \includegraphics[width=1.\columnwidth]{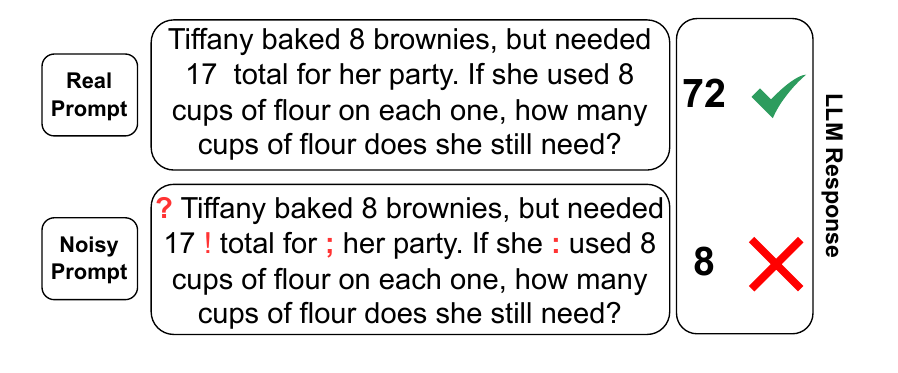}
    \caption{Noisy context breaks the LLM's capability to give the right answer.}
    \label{fig:ArithmAttack-sample}
\end{figure}

In this paper, we further investigate the math problem-solving robustness of LLMs to a different set of changes that take the form of noisy context containing a variety of punctuation marks. Given that none of the words are changed when new punctuation marks are inserted into the input text, the semantic similarity of the perturbed sentence remains unchanged. The key research question for this study is: \textit{How do LLMs respond to noise attacks consisting of random punctuation marks in the context of math problem-solving?} Figure \ref{fig:ArithmAttack-sample} shows an example of an LLM response under ArithmAttack, where the model behaves erratically when it sees a noisy context whereas it answers the question in the clean prompt correctly.

Inspired by the AEDA (An Easier Data Augmentation) method \cite{karimi-etal-2021-aeda-easier}, we propose ArithmAttack to assess the robustness of eight LLMs (i.e. two Llama models \cite{dubey2024llama}, two Mistral models \cite{jiang2023mistral7b}, Zephyr \cite{tunstall2023zephyr}, Gemma2 \cite{team2024gemma}, Qwen2.5 \cite{yang2024qwen2}, and DeepSeek \cite{guo2025deepseek}) to noisy data. Similarly to AEDA, we introduce this noise by randomly inserting punctuation marks into the context of math problems from two math datasets, namely GSM8K \cite{cobbe2021gsm8k} and MultiArith \cite{roy2015solving}. We then evaluate how these models perform under different noise levels, with the noise affecting 10\%, 30\%, and 50\% of the sentence length (based on the number of words).

Our contributions are twofold: 1) We propose ArithmAttack which produces noisy contexts containing random punctuation marks to assess the robustness of LLMs in math problem-solving. 2) We evaluate eight LLMs, with parameter counts of 1.5B, 2B, 7B, and 8B on math datasets and observe that all the studied models show growing vulnerability to ArithmAttack as the amount of noise increases.

\section{Related Work}
Noise insertion has been shown to be effective in deteriorating the performance of encoder models in various tasks such as toxic text classification \cite{hosseini2017deceiving, eger2020hero}, sentiment analysis \cite{formento2021special}, and natural language inference \cite{formento2023using}.

In the context of math problem solving, Large Language Models (LLMs) have been shown to be vulnerable to a variety of changes in the input context, including  
typographical errors \cite{gan2024reasoning}, word replacement \cite{zhou2024mathattack}, gibberish or irrelevant context inclusion \cite{shi2023large}, and semantic perturbations \cite{zhu2023promptbench}.  
\citet{gan2024reasoning} propose an adversarial typo attack that breaks the reasoning process of LLMs. Instead of modifying characters, \citet{zhou2024mathattack} propose a dataset, called RobustMath, where they replace words with their synonyms to evaluate the robustness of large language models. Similarly, \citet{li2024gsm} propose GSM-plus dataset, based on GSM8K, modified with a variety of mathematical perturbations such as distractor insertion and arithmetic variation. 
In the study by \citet{zhu2023promptbench}, the authors employ different types of textual attacks on prompts, including character, word, sentence, and semantic attacks. In contrast, \citet{xie2024adversarial} propose to modify the numeric values in the questions using abstract syntax trees resulting in examples that fool the LLMs. 

While the literature mainly concentrates on modifying the lexical or semantic content of the prompts, we aim to keep the contextual information intact and instead focus on the model behavior changes in reasoning when encountering punctuation noise. In addition, an advantage of our method is that it is extremely straightforward to implement, and as we show in the results section, it is also effective in degrading the performance of LLMs in math problem-solving.

\section{Experiments}
To carry out our experiments, we use two well-known math datasets and eight LLMs. 

\subsection{Datasets}
\noindent
\textbf{GSM8K} \cite{cobbe2021gsm8k} contains 8.5K high-quality, linguistically diverse grade school math
word problems. The test set contains 1.32k data points on which we do our experiments. This dataset provides a variety of arithmetic and logical questions typical of
middle school education, making it ideal for testing comprehension and problem-solving
capabilities of LLMs under noisy conditions. 

\noindent
\textbf{MultiArith} \cite{roy2015solving} offers a broad examination of language model performance across multiple arithmetic problem types and complexities. The test set contains 180 data points on which we do our experiments. It serves as a crucial benchmark for understanding how contextual noise impacts the model's ability to parse and solve mathematical questions. 

\subsection{Models}
To study a variety of language models and at the same time observe our computational budget, we opted for eight widely utilized LLMs that have been trained by different companies. These models are \textbf{Mistral-7B-Instruct-v0.2} \cite{jiang2023mistral7b}, \textbf{Mathstral-7b-v0.1} \cite{jiang2023mistral7b}, \textbf{Llama-3-8B-Instruct} and \textbf{Llama-3.1-8B-Instruct} \cite{dubey2024llama}, \textbf{Gemma-2-2b-it} \cite{team2024gemma}, \textbf{Zephyr-7b-beta} \cite{tunstall2023zephyr}, \textbf{Qwen2.5-1.5B-Instruct} \cite{yang2024qwen2}, and \textbf{DeepSeek-R1-Distill-Llama-8B} \cite{guo2025deepseek}. Throughout this paper, we will refer to these models as Mistral, Mathstral, Llama3, Llama3.1, Gemma2, Zephyr, Qwen2.5, and DeepSeek respectively.

\section{Methodology}
To obtain the responses from LLMs, we use the Zero-Shot CoT \cite{kojima2022large} prompting, using the following prompt: 

\begin{mytheo}{}{}
    \texttt{Think step by step through the following problem and clearly show each step of your reasoning. Ensure the final answer is indicated by ending with \{The final answer is\}}.
\end{mytheo}

\begin{table}
\setlength{\tabcolsep}{3.5pt}
\centering
\begin{tabularx}{\columnwidth}{lccccHHrr} 
\toprule
Models & \makecell{Clean \\ Acc (\%)} & \multicolumn{3}{c}{ \makecell{Punctuation \\ Percentage}} & \makecell{Clean Incorrect \\ Extraction (\%)} & \makecell{Clean False \\ Negative (\%)} & ASR \\
\cline{3-5}
& & 10 & 30 & 50 & &&\\
\midrule
Mistral    & 42.07 & 41.62 & 37.75 & 36.39 & 18.00  & 9.00   & 39.69 \\
Mathstral     & 77.63 & 75.51 & 71.34 & 70.65 & 1.00   & 1.00   & 19.81    \\
Llama3    & 75.43 & 73.31 & 73.08 & 72.17 & 1.00   & 0.00   & 16.04     \\
Llama3.1     & \textbf{82.25} & \textbf{81.04} & \textbf{78.84} & \textbf{77.02} & 0.00   & 0.00   & \textbf{12.53}      \\
Gemma2    & 49.65 & 45.10 & 36.46 & 35.63 & 8.00   & 5.00   & 41.82     \\
Zephyr     & 23.27 & 18.04 & 18.04 & 10.08 & 31.00   & 29.00   & 74.80     \\
Qwen2.5 & 61.10 & 56.02 & 52.69 & 49.35 & 6.00   & 5.00   & 31.59      \\
DeepSeek & 73.76 & 73.76 & 70.43 & 67.24 &    &    & 20.46      \\
\bottomrule
\end{tabularx}
\caption{Results for GSM8K dataset (numbers are in percentages). The performance for all models drops under ArithmAttack. Llama3.1 has the top performance under all levels of noise.}
\label{table:performance_gsm}
\end{table}

\subsection{Noisy Dataset Creation}
Once satisfactory results were achieved with clean datasets, we proceeded to test the models on noisy data. 
For the introduction of noise, we follow a similar approach to \citet{karimi-etal-2021-aeda-easier}, by altering the hyperparameters in the logic. In their study, they insert the punctuation marks by randomly choosing a number between 1 and one-third of the length of the sequence which indicates how many insertions will be carried out. But in our case, instead of randomly choosing the number of insertions, we fix it to be 10\%, 30\%, and 50\% of the total length of the sentence but still choose random positions to insert the noise. We employed six types of punctuation marks: \texttt{\{".", ',', '!', '?', ';', ':'\}}.

\subsection{ASR and Similarity Calculation}
We evaluate the models with their performance accuracy against noisy input and \textbf{Attack Success Rate (ASR)}. ASR \cite{wang2adversarial} measures how effective an adversarial attack is on a model. Specifically, it looks at how often the model's predictions are changed incorrectly after the adversarial attack. In this study, the average ASR has been taken for every model with 10\%, 30\% and 50\% noisy dataset's responses with the help of Formula \ref{formula:asr}:  \\
\begin{equation}\label{formula:asr}
ASR = \frac{\sum_{(x, y) \in D} \mathbb{I}\left[f(A(x)) \neq y\right]}{\sum_{(x, y) \in D} \mathbb{I}\left[f(x) = y\right]}
\end{equation}

In other words, ASR is the ratio of changed answers after attack to previously correct answers produced by the LLM.

We also calculate the similarity of the perturbed samples to the original ones. \textbf{Similarity} represents the average semantic similarity between two contexts. Given that our method does not alter the words in the sentence, the resulting samples after applying ArithmAttack are scored 100 percent similar to the original samples using Universal Sentence Encoder \cite{cer2018universal} as the scorer. This indicates that our noise insertion attack does not impose any semantic shifts on the input text. 

\begin{table}
\setlength{\tabcolsep}{3.5pt}
\centering
\begin{tabularx}{\columnwidth}{lccccHHrr} 
\toprule

Models & \makecell{Clean \\ Acc (\%)} & \multicolumn{3}{c}{ \makecell{Punctuation \\ Percentage}} & \makecell{Clean Incorrect \\ Extraction (\%)} & \makecell{Clean False \\ Negative (\%)} & ASR \\
\cline{3-5}
& & 10 & 30 & 50 & &&\\
\midrule
Mistral    & 73.88 & 72.77 & 71.11 & 65.55 & 4.44  & 3.33   & 23.66     \\
Mathstral        & 96.11 & 92.77 & 86.11 & 87.22 & 1.11   & 0.00   & 9.47   \\
Llama3    & 95.00 & 92.77 & \underline{91.66} & \underline{88.33} & 1.11   & 0.00   & \textbf{7.79}     \\
Llama3.1    &  \textbf{99.44} & \textbf{94.44} & \underline{91.66} & 83.88 & 0.55   & 0.55   & 9.67      \\
Gemma2     & 89.44 & 82.77 & 78.88 & 72.22 & 3.8   & 1.66   & 19.45     \\
Zephyr    & 37.22 & 22.22 & 16.11 & 12.77 & 36.66   & 23.88   & 77.10  \\
Qwen2.5 & 97.22 & \textbf{94.44} & 85.55 & 83.88 & 0.55   & 0.00   & 11.04    \\
DeepSeek & 93.88 & 90.00 & \textbf{92.77} & \textbf{88.88} &    &    & \underline{8.28}    \\
\bottomrule
\end{tabularx}
\caption{Results for MultiArith dataset (numbers are in percentages). The performance for all models drops under ArithmAttack. Llama3 has the lowest drop, making it more robust than others. }
\label{table:performance_ma}
\end{table}

\begin{figure*}[t]
\centering
\subfigure{\includegraphics[width=0.36\textwidth]{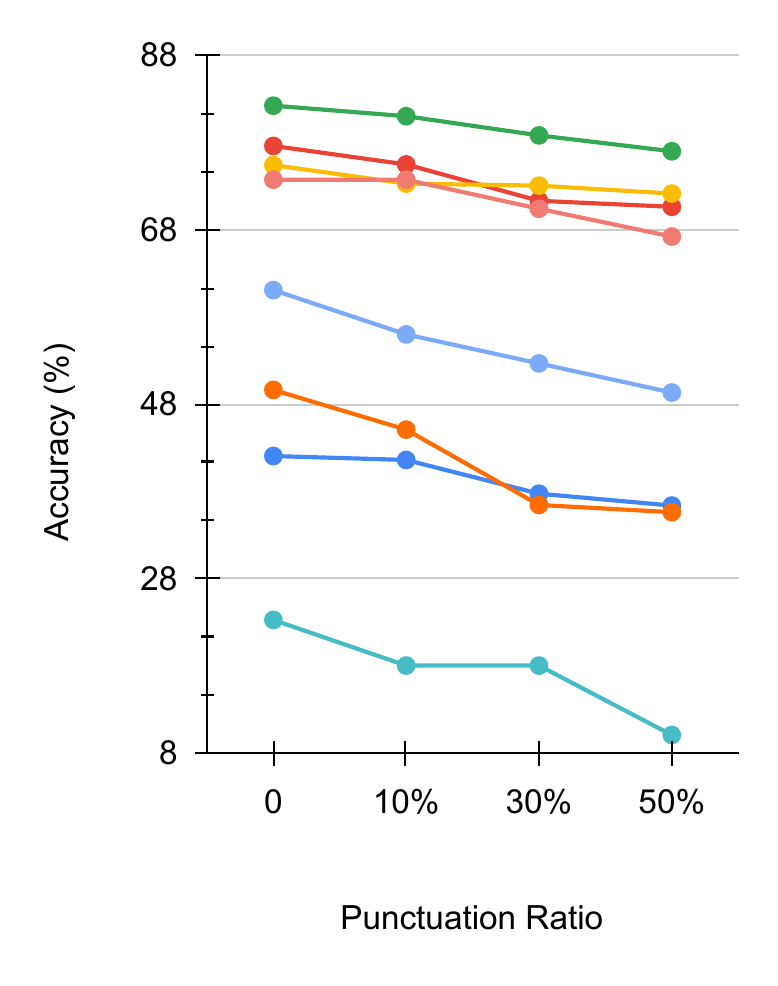}}
\hspace{10mm}
\subfigure{\includegraphics[width=0.48\textwidth]{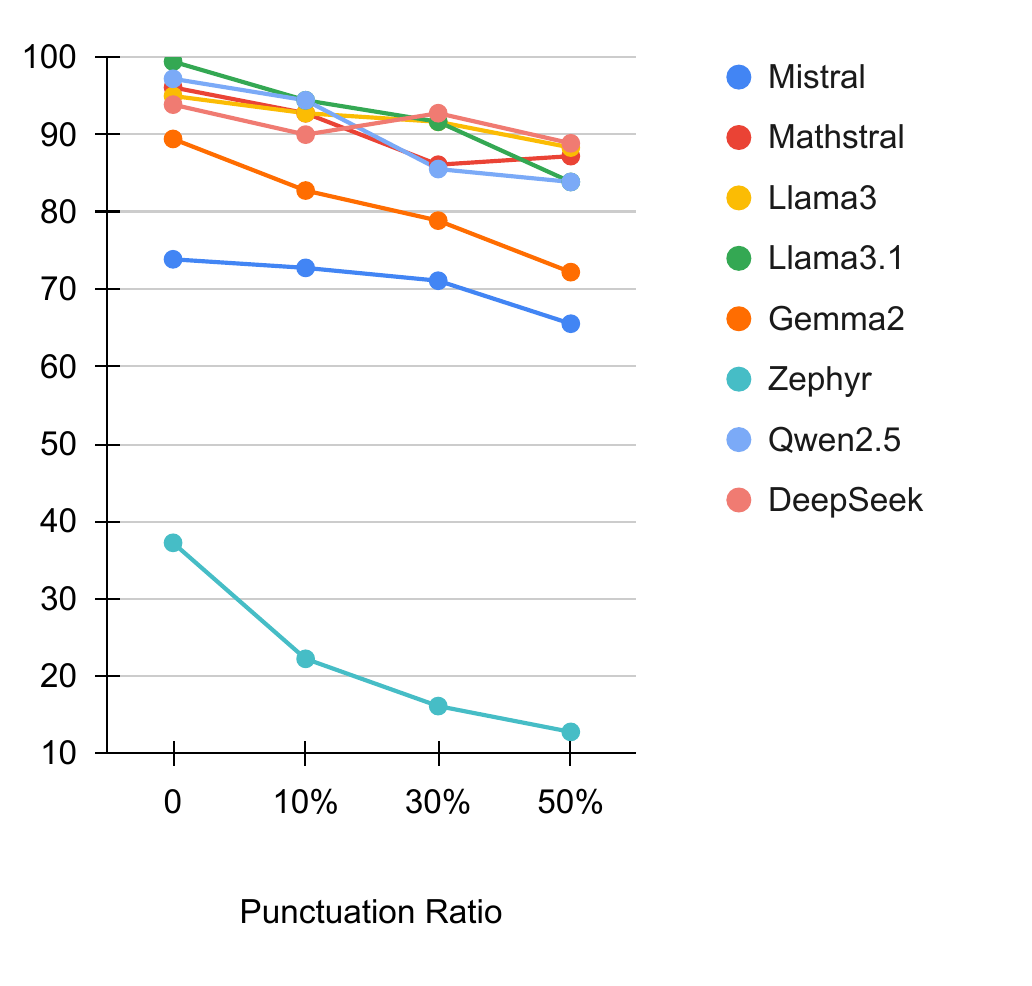}}
\caption{Accuracy of the studied models on different levels of noise for GSM8K (left) and MultiArith (right) datasets. Llama models show the highest robustness as well as performance. }
\label{fig:performance_drop}
\end{figure*}

\begin{figure}[t]
    \centering
    \includegraphics[width=\linewidth]{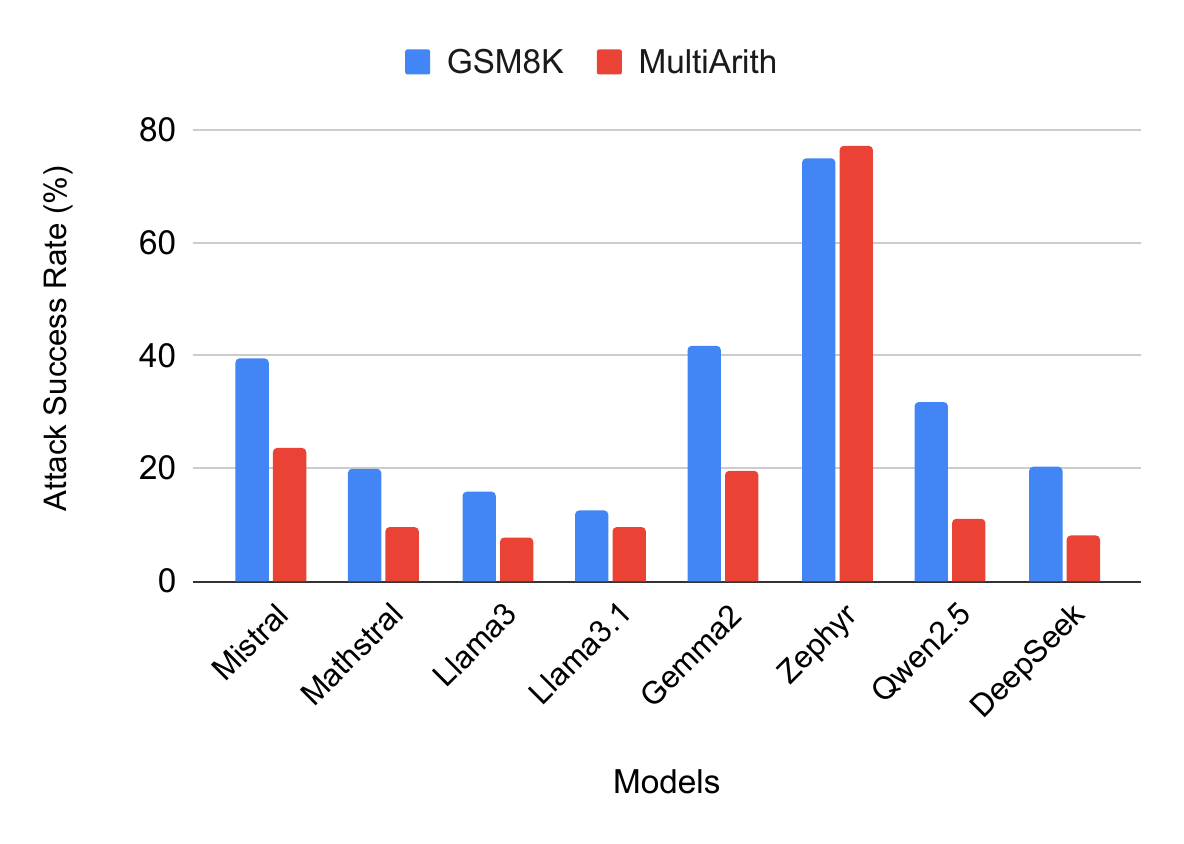}
    \caption{Comparing the attack success rates on the studied models for GSM8K and MultiArith datasets (lower is better for model robustness). Llama models are more robust than others under ArithmAttack.}
    \label{fig:asr}
\end{figure}

\section{Results and Analysis}
As shown in Tables \ref{table:performance_gsm} and \ref{table:performance_ma}, Llama3.1 outperforms other models across both datasets in the majority of the cases. It achieves the highest accuracies in both clean and noise-affected settings (except in 30\% and 50\% noisy data of the MultiArith dataset where DeepSeek in the former and both Llama3 and DeepSeek in the latter have higher accuracies). This makes it the most reliable model for handling mathematical problems under noisy input conditions. In terms of the ASR score, Llama3.1 has the lowest score in GSM8K and Llama3 in MultiArith, indicating that Llama models are more robust to noise than other studied models with the exception of DeepSeek only in MultiArith dataset showing comparable robustness. In addition, the Mathstral model compared to Mistral exhibits more robustness which can be attributed to its higher mathematical understanding.
In contrast, Zephyr was the lowest-performing model, exhibiting low clean accuracy and suffering a significant decline in performance as noise was introduced. Its high ASR score makes it unsuitable for tasks involving noisy data, reflecting poor robustness.

Figure \ref{fig:performance_drop} shows the relationship between the model’s accuracy and the noise present in the datasets. For both datasets, as the percentage of noise in the data increases, the accuracy decreases. This indicates that these models are not robust against noise in the data. This also provides a future direction for improving these models and making them more robust to noise.

Across all models except for Zephyr, the impact of noise was more pronounced in the GSM8K dataset than in MultiArith, with a larger drop in accuracy as the noise levels increased (Figure \ref{fig:asr}). In manual inspection, we found out that the GSM8K dataset was more difficult to solve than the MultiArith dataset. This suggests that the models may struggle more with noise in math datasets with more difficulty.

\begin{table}
    \centering
    \begin{tabular}{lc c}
        & \multicolumn{2}{c}{Miss Rate (\%)} \\
        \cline{2-3}
       Model  & GSM8K & MultiArith \\
        \toprule
       Mistral  & 9.0 & 1.1 \\
       Mathstral  & 0.0 & 1.1\\
       Llama3  & 1.0 & 1.1 \\
       Llama3.1  & 0.0 & 0.0\\
       Gemma2  & 3.0 & 2.2 \\
       Zephyr  & 2.0 & 12.8\\
       Qwen2.5  & 1.0 & 0.5\\
       DeepSeek  & 4.0 & 0.0\\
        \bottomrule
    \end{tabular}
    \caption{Miss rate of the models in answer extraction}
    \label{tab:miss_rate}
\end{table}

\paragraph{Answer Extraction Accuracy}
To evaluate the accuracy of the models, we developed a script to extract answers from the LLM responses. The extraction process underwent multiple iterations, as it needed to accurately extract the answer and compare it with the ground truth. However, even with the final prompt, we observed a couple of inconsistencies in the answer extraction. Therefore, we went through outputs manually to estimate the \textit{miss rate} (i.e. the rate with which the correct answer is not extracted). In manual inspection, we evaluated the entire responses for the MultiArith dataset and the first 100 responses for the GSM8K dataset from all the models except for the DeepSeek model. For this model (due to time and labor constraints), we evaluated the first 50 samples from each dataset. Table \ref{tab:miss_rate} shows that the miss rate is minimal for most of the models. In the cases of Mistral (for GSM8K) and Zephyr (for MultiArith), the miss rates can be significant. While this can be an indication of lower ability in following instructions in these models, considering the gap in the performance and ASR scores, this does not affect the observed trends. 

\section{Conclusions and Future Work}
We evaluated how well different language models handle mathematical problem-solving tasks in both clean and noisy conditions. Our results indicate that all studied models can be vulnerable to extra noise with varying degrees, with Llama models being the highest-performing and the most robust model in the majority of the experiments. In addition, comparing the two models of Mathstral and Mistral from the same family, the one with mathematical knowledge exhibited more robustness to noise. Lastly, the findings revealed that more complex datasets such as GSM8K can become more difficult to understand as they become noisier. Future research can include datasets beyond GSM8K and MultiArith as well as other reasoning tasks such as logical and causal reasoning, which could provide deeper insights into the models' robustness in different scenarios. Further experimentation with different types of noise could also help enhance our understanding of the latent vulnerabilities in LLMs. Finally, explaining why ArithmAttack can break the reasoning flow of LLMs would be another valuable follow-up to this work. 

\section{Limitations}
To make the questions noisy, we have opted for one type of noise which is irregular use of punctuation marks. While some of the other noise types such as spelling and typographical errors have been studied in the literature (mentioned in related work), there are other types such as grammatical errors, wrong abbreviations, and acronyms that we have not explored. In addition, to observe our computational budget, we have utilized only two math datasets and eight LLMs. For a more comprehensive experimentation, one can experiment with other available math datasets and a larger number of LLMs. 

\section*{Acknowledgments}
This work was partially supported by the AISafety Project, funded by the BMBF under the grant proposal 05D23PD1, and by the state of North Rhine-Westphalia as part of the Lamarr Institute for Machine Learning and Artificial Intelligence. 
We would also like to thank the reviewers for their invaluable comments, which helped strengthen the quality of this work.
\bibliography{ref}

\end{document}